%% file: main.tex
\documentclass[letterpaper]{article} 
\usepackage{aaai2026}  
\usepackage{times}  
\usepackage{helvet}  
\usepackage{courier}  
\usepackage[hyphens]{url}  
\usepackage{graphicx} 
\usepackage{natbib}  
\usepackage{caption} 
\usepackage{algorithm}
\usepackage{algorithmic}

\usepackage{subcaption} 
\usepackage{enumitem}
\usepackage{booktabs} 
\usepackage{caption}  
\usepackage{amsmath}
\usepackage{amsfonts}

\usepackage{times}
\usepackage{helvet}
\usepackage{courier}
\usepackage{xcolor}

\usepackage{newfloat}
\usepackage{listings}
\DeclareCaptionStyle{ruled}{labelfont=normalfont,labelsep=colon,strut=off} 
\floatstyle{ruled}
\newfloat{listing}{tb}{lst}{}
\floatname{listing}{Listing}
\title{Coordinated Humanoid Robot Locomotion with Symmetry Equivariant Reinforcement Learning Policy}
\author{
    Buqing Nie\equalcontrib\textsuperscript{\rm 1}, 
    Yang Zhang\equalcontrib\textsuperscript{\rm 1},
    Rongjun Jin\equalcontrib\textsuperscript{\rm 1},
    Zhanxiang Cao\textsuperscript{\rm 1,\rm 2}, \\
    Huangxuan Lin\textsuperscript{\rm 1}, 
    Xiaokang Yang\textsuperscript{\rm 1},
    Yue Gao\thanks{Corresponding author.}\textsuperscript{\rm 1,\rm 2}
}
\affiliations{
    \textsuperscript{\rm 1}MoE Key Lab of Artificial Intelligence and AI Institute,
    Shanghai Jiao Tong University, Shanghai, China \\
    \textsuperscript{\rm 2} Shanghai Innovation Institute, Shanghai,  China

    \{niebuqing,zhangyang-sjtu-2022,k4rock,caozx1110,b3mylq,xkyang,yuegao\}@sjtu.edu.cn
}

\usepackage{bibentry}

\begin{document}

\maketitle

\begin{abstract}
The human nervous system exhibits bilateral symmetry, enabling coordinated and balanced movements.
However, existing Deep Reinforcement Learning (DRL) methods for humanoid robots neglect morphological symmetry of the robot, leading to uncoordinated and suboptimal behaviors. 
Inspired by human motor control, we propose Symmetry Equivariant Policy (SE-Policy), a new DRL framework that embeds strict symmetry equivariance in the actor and symmetry invariance in the critic without additional hyperparameters. 
SE-Policy enforces consistent behaviors across symmetric observations, producing temporally and spatially coordinated motions with higher task performance.
Extensive experiments on velocity tracking tasks, conducted in both simulation and real-world deployment with the Unitree G1 humanoid robot, demonstrate that SE-Policy improves tracking accuracy by up to 40\% compared to state-of-the-art baselines, while achieving superior spatial-temporal coordination. 
These results demonstrate the effectiveness of SE-Policy and its broad applicability to humanoid robots.
\end{abstract}
\section{Introduction}

Recently, Deep Reinforcement Learning (DRL) has achieved remarkable achievements in robotics control tasks, including quadruped robots~\cite{nahrendra2023dreamwaq}, manipulators~\cite{singh2019end}, bipedal robots~\cite{li2025reinforcement}, and humanoid robots~\cite{he2024omniho}.
DRL enables robots to acquire agile skills through interactions in simulation environments autonomously without extensive domain knowledge~\cite{radosavovic2024real}.

However, existing DRL policies are typically black-box in nature, failing to leverage the inherent skill sharing offered by the symmetric morphology property of the humanoid robot ~\cite{zinkevich2001symmetry,van2020mdp,panangaden2024policy}. 
Such policies exhibit inconsistent reactions to symmetrically equivalent observations, such as unequal movement styles of symmetric joints during locomotion, revealing a limited understanding to the robot morphology~\cite{apraez2025morphological,ding2024breaking}.
Such oversight leads to asymmetric and uncoordinated behaviors, consequently yielding unnatural and suboptimal policies that negatively impact user experience and diminish task performance~\cite{su2024leveraging,mittal2024symmetry}.

In order to tackle this problem, previous works propose various novel methods to incorporate  morphological symmetry into the training framework~\cite{su2024leveraging,mittal2024symmetry,abdolhosseini2019learning,wang2022robot}.  
Some research improves symmetry performance from \emph{temporal} perspective, i.e. encourages periodicity of robot motions~\cite{lin2020invariant,lee2020learning,gu2024humanoid,ding2024breaking}.
For instance, some prior works~\cite{gu2024humanoid,ding2024breaking} introduce periodic phase signals into the observation and reward design, encouraging the periodical gait motions of the policy.
Lee et al.~\cite{lee2020learning} design a novel action space based on Central Pattern Generators~\cite{bellegarda2022cpg} to describe temporally symmetric motions.

Besides, other works aims to address this problem
from a \emph{spatial} perspective, i.e. output equivariant actions under symmetric states~\cite{mittal2024symmetry,wang2022robot,su2024leveraging}.
Some works augment collected transitions with their symmetric copies to induce equivariance and invariance for the actor and critic  correspondingly, which is effective in quadruped locomotion~\cite{abdolhosseini2019learning,mittal2024symmetry} and manipulation~\cite{lin2020invariant}. 
Another approach is introducing regularization into the optimization objective of the actor and critic training, which has been widely utilized in RL-based robot methods~\cite{gu2024humanoid,abreu2025addressing,ben2025homie,long2024learning,xue2025unified}.
Prior works also explore enforcing equivariance on RL through introducing hard constraints on neural network-based policy architectures~\cite{mondal2020group,mondal2022eqr}.
This approach is effective on classic control tasks~\cite{van2020mdp,rezaei2022continuous}, quadruped control~\cite{su2024leveraging}, and manipulations~\cite{wang2022robot,wang2022so}.

Despite previous novel methods, the optimal approach to integrating symmetry equivariance into DRL-based robot policies still remains underexplored, particularly on real humanoid robots, which demand high agility and robustness in their motions~\cite{zhuang2024humanoid,zhang2025natural}.
Loosely equivariant methods, such as temporal symmetric policies and data augmentation, show moderate performance due to insufficient policy constraints, and are commonly employed as auxiliary techniques in robot tasks~\cite{long2024learning,xue2025unified}.
Loss regularization methods introduce additional hyperparameters requiring delicate tuning, and these terms may impede the optimization process of policies~\cite{mittal2024symmetry,su2024leveraging}.
Strict equivariant methods show promise predominantly in simulated classic control tasks, where their effectiveness on robots, especially real humanoid robots, remains under-explored~\cite{van2020mdp,rezaei2022continuous,wang2022so}.

In this work, we propose a new DRL-based method for humanoid robot control tasks, called \textbf{S}ymmetry \textbf{E}quivariant \textbf{Policy} \textbf{(SE-Policy)}.
This method induces strict symmetry equivariance and invariance into the network architectures of the actor and critic respectively.
This leads to more coordinated and natural motions without introducing additional hyperparameters. 
Experiments are conducted in both simulation and on a real humanoid robot through sim-to-real, demonstrating superior locomotion performance compared to previous methods.

The main contributions of this work can be summarized as follows:
\begin{itemize}
    \item We propose a new method SE-Policy for humanoid robot control tasks, which incorporates strict symmetry equivariance property into actor-critic architecture without additional hyper-parameters.  
    \item SE-Policy generates symmetric and natural motions, achieving higher control performance on tracking error and coordination compared to previous methods.
    \item Experiments are conducted in both the simulation environments and on the real humanoid robot via sim-to-real, demonstrating superior performance of our method.
\end{itemize}

\section{Related Works}

\subsection{Equivariant Reinforcement Learning}

Equivariance property is actively studied in DRL studies to improve sample efficiency and policy performance~\cite{panangaden2024policy,rezaei2022continuous,wang2022so}.
Zinkevich et al.~\cite{zinkevich2001symmetry} formulate symmetric MDP and prove the equivariance/invariance for actor/critic in RL. 
Some works implement equivariant policy through data augmentation~\cite{lin2020invariant,corrado2024understanding}.
For example, Luo et al.~\cite{luo2024reinforcement} propose to conduct data augmentation based on Euclidean symmetries, achieving improved data efficiency.
Park et al.~\cite{park2025approximate} propose a novel equivariant RL method to tackle MDP with approximate equivariant structures.
Some other works, such as Eqr~\cite{mondal2022eqr}, utilize policy equivariance to improve representation learning, achieving higher sample efficiency.
In addition, equivariance has demonstrated excellent performance across various task domains, including discrete action spaces~\cite{van2020mdp,mondal2020group}, continuous control~\cite{rezaei2022continuous,yarats2021improving}, image observations~\cite{nguyen2023equivariant}, state observations~\cite{wang2022so,panangaden2024policy}, and multi-agent RL domains~\cite{chen2024rm,bousias2025symmetries}.
In this work, we integrate equivariance into RL policy for humanoid robots, investigating its influence on humanoid robot control tasks.

\subsection{Symmetry Equivariant Robot Policy}

Soft equivariant policy has shown effectiveness in various robot tasks, including path planning~\cite{theile2024equivariant}, robot manipulation~\cite{wang2022robot}, grasping~\cite{hu2025orbitgrasp}, quadruped locomotion~\cite{su2024leveraging}, and humanoid imitation learning~\cite{long2024learning}.
Policy equivariance is effective to increase sample efficiency~\cite{mittal2024symmetry}, enhance motion coordination~\cite{apraez2025morphological}, and improve policy performance~\cite{ben2025homie}.
Current works enhance policy equivariance mainly through reward shaping~\cite{ding2024breaking}, data augmentation~\cite{mittal2024symmetry}, and loss regularization~\cite{xue2025unified}.
These methods induce soft constraints on policies, resulting in loose equivariance with moderate performance.
In addition, some methods such as loss regularization may hinder the training of the RL algorithm~\cite{mittal2024symmetry,su2024leveraging}.
Moreover, the influence of equivariance on humanoid robot has not been fully investigated.

\section{Method}

\begin{figure*}[t]
\centering
\includegraphics[width=0.99\linewidth]{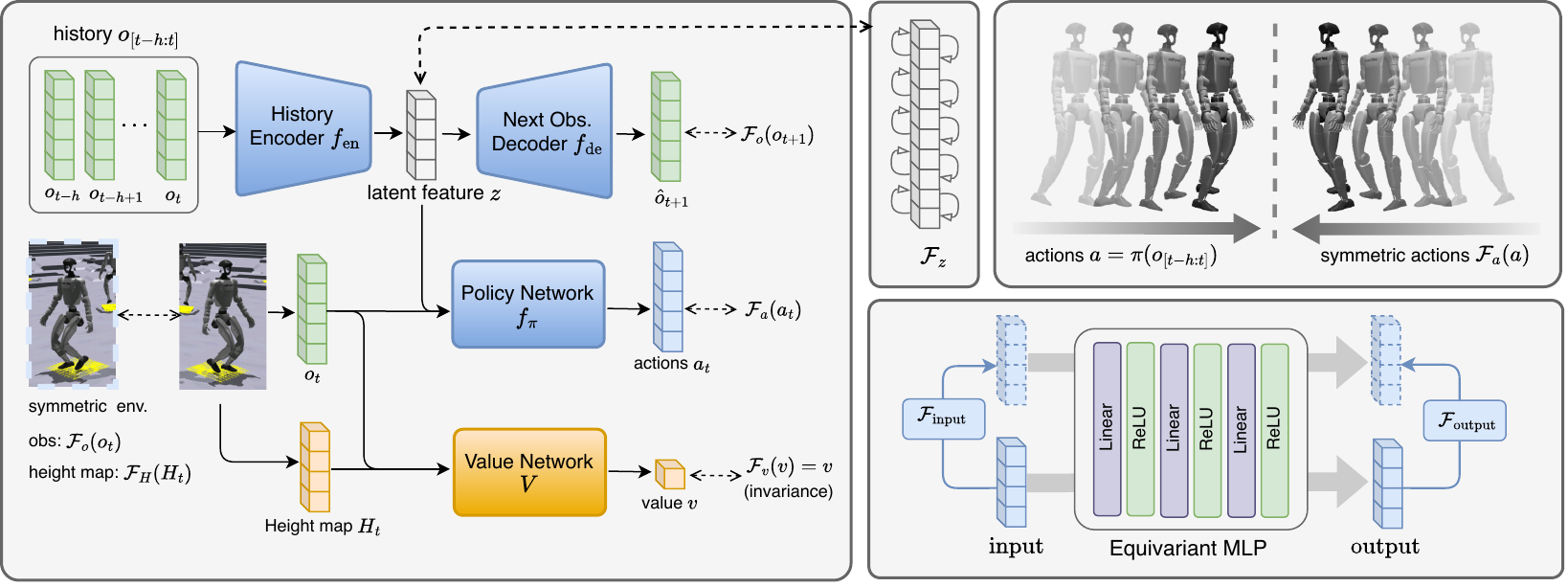} 
\caption{
The overall architecture of \emph{SE-Policy}. \textbf{(a) 
 Left}: the architecture of the actor and critic model. 
 \textbf{(b) upper right:} the visualization of $\mathcal{F}_z$, i.e. the symmetric transformation of $z$.
 The visualization of humanoid robot motions and corresponding symmetric motions.
 \textbf{(b) bottom right:} the description of equivariant MLP, which is widely utilized in this work.
}
\label{fig:whole_arch}
\end{figure*}

\subsection{Problem Formulation}
\label{sec:problem_formulation}

The interaction between the humanoid robot and the environment can be formulated as 
Markov decision process (MDP)
$\mathcal{M} = \langle\mathcal{S}, \mathcal{O}, \mathcal{A}, {P}, {R}, \gamma\rangle$, where $s_t \in\mathcal{S}$, $o_t \in\mathcal{O}$, and $a_t \in\mathcal{A}$ denote the robot state, observation, and action at $t$-th timestep respectively.
$P$ denotes the transition probability of the environment.
$R:\mathcal{S}\times \mathcal{A}\to \mathbb{R}$ denotes the reward function, and $\gamma \in [0,1)$ denotes the discount factor.
The agent takes action $a_t$ according to its policy $\pi$ at each step $t$ given the observation, i.e. $a_t \sim \pi$.
The objective of the robot is to maximize the cumulative episode return, which is formulated as follows:
\begin{equation}
\label{eq:optimize_pi_objective}
\pi_{}^{*} = \mathop{\arg\max}\limits_{\pi}\mathbb{E}_{a_t\sim\pi,s_{t+1}\sim \mathcal{P}}\left[\sum_{t}\gamma^{t}R(s_{t},a_{t})\right].
\end{equation}
In order to conduct policy evaluation, we can define the state-action value function 
\begin{equation}
\label{eq:Q_func}
    Q(s_t,a_t) = R(s_t, a_t) + \gamma \mathbb{E}_{\pi,P}\left[ \sum_{t'=t}^{\infty} \gamma^{t'-t}R(s_{t'}, a_{t'}) \right]
\end{equation} 
as the discounted return starting from $s_t$, given that $a_t$ is taken and then $\pi$ is followed.
The value function $V(s_t) = \mathbb{E}_{a_t\sim\pi} \left[Q(s_t,a_t)\right]$ denotes the discounted return starting from $s_t$ following $\pi$.

In this work, we utilize SE-Policy for the velocity tracking task of Unitree G1, a versatile humanoid robot frequently used in robotics research~\cite{unitreeHumanoidRobot}.
Note that this method is a general training framework, thus can be applied to diverse control tasks and humanoid robots that possess morphological symmetry.

\subsection{Symmetry Equivariant Policy}

\subsubsection{Observation Space}
In this work, the observation of the humanoid robot $o_t \in \mathbb{R}^{96}$ is described in Table~\ref{tab:obs_and_action_transformation}, i.e.
\begin{equation}
\boldsymbol{o}_t=\begin{bmatrix}\boldsymbol{\omega}&\boldsymbol{g}&\boldsymbol{c}&\boldsymbol{\theta}&\boldsymbol{v}&\boldsymbol{a}_{t-1}&\boldsymbol{\Phi}\end{bmatrix}^T,
\end{equation}
where the user-provided velocity command $c = (c_{x}, c_{y}, c_{\omega})$ specifies the desired linear velocities along the $x$ and $y$ axes $(c_x, c_y)$, and the angular velocity $c_{\omega}$.
The phase input $\Phi$ is a sinusoidal clock signal, which is employed to generate reference motion cycles.
This design is widely used in RL-based control works to improve temporal symmetry (periodicity) of the gait. 
The height map $H$ describes robot-centric terrain information and is utilized for critic training only.

\subsubsection{Action Space}

This work employs position control for the humanoid robot.
Accordingly, the action space $|\mathcal{A}|=27$ denotes the target joint positions of each joint.
The detailed description of the observation and action space is given in Appendix 1.3.

\begin{table*}[tbp]
    \centering
    \begin{tabular}{llcc} 
        \toprule
        \textbf{Component} & \textbf{Dim} & \textbf{Description} & \textbf{Symmetry Transformation $\mathcal{F}$} \\
        \midrule
    
        Base angular velocity $\boldsymbol{\omega}$  & $o_t^{1:3}$ & $(\omega_{x}, \omega_{y}, \omega_{z})$ & $(-\omega_{x}, \omega_{y}, -\omega_{z})$\\
        Projected gravity $\boldsymbol{g}$  & $o_t^{4:6}$ & $(g_{x}, g_{y}, g_{z})$ & $(g_{x}, -g_{y}, g_{z})$  \\
        Velocity commands $\boldsymbol{c}$ & $o_t^{7:10}$ & $(c_{x}, c_{y}, c_{\omega})$ & $(c_{x}, -c_{y}, -c_{\omega})$\\
        Joint positions $\boldsymbol{\theta}$ & $o_t^{11:37}$ & $(\theta_{\operatorname{left}}^{\operatorname{arm}}, \theta_{\operatorname{right}}^{\operatorname{arm}}, \theta_{\operatorname{left}}^{\operatorname{leg}}, \theta_{\operatorname{right}}^{\operatorname{leg}}, \theta_{\operatorname{waist}})$ & $(-\theta_{\operatorname{right}}^{\operatorname{arm}}, -\theta_{\operatorname{left}}^{\operatorname{arm}}, -\theta_{\operatorname{right}}^{\operatorname{leg}}, -\theta_{\operatorname{left}}^{\operatorname{leg}}, -\theta_{\operatorname{waist}})$\\
        Joint velocities $\boldsymbol{v}$ & $o_t^{38:65}$ & $(v_{\operatorname{left}}^{\operatorname{arm}}, v_{\operatorname{right}}^{\operatorname{arm}}, v_{\operatorname{left}}^{\operatorname{leg}}, v_{\operatorname{right}}^{\operatorname{leg}}, v_{\operatorname{waist}})$ & $(-v_{\operatorname{right}}^{\operatorname{arm}}, -v_{\operatorname{left}}^{\operatorname{arm}}, -v_{\operatorname{right}}^{\operatorname{leg}}, -v_{\operatorname{left}}^{\operatorname{leg}}, -v_{\operatorname{waist}})$\\
        Previous action $\boldsymbol{a_{t-1}}$ & $o_t^{66:93}$ & $(a_{\operatorname{left}}^{\operatorname{arm}}, a_{\operatorname{right}}^{\operatorname{arm}}, a_{\operatorname{left}}^{\operatorname{leg}}, a_{\operatorname{right}}^{\operatorname{leg}}, a_{\operatorname{waist}})$ & $(-a_{\operatorname{right}}^{\operatorname{arm}}, -a_{\operatorname{left}}^{\operatorname{arm}}, -a_{\operatorname{right}}^{\operatorname{leg}}, -a_{\operatorname{left}}^{\operatorname{leg}}, -a_{\operatorname{waist}})$\\
        Phase input $\boldsymbol{\Phi}$  & $o_t^{94:96}$ & $(\Phi_{sin}, \Phi_{cos})$ & $(-\Phi_{sin}, -\Phi_{cos})$  \\
        \midrule 
        Height map $\boldsymbol{H}$  & $H^{1:187}_t$ & $(H_{\operatorname{left}}, H_{\operatorname{middle}}, H_{\operatorname{right}})$ & $(H_{\operatorname{right}}, H_{\operatorname{middle}}, H_{\operatorname{left}})$ \\
        \midrule
        Action $\boldsymbol{a_{t}}$ & $a_t^{1:27}$ & $(a_{\operatorname{left}}^{\operatorname{arm}}, a_{\operatorname{right}}^{\operatorname{arm}}, a_{\operatorname{left}}^{\operatorname{leg}}, a_{\operatorname{right}}^{\operatorname{leg}}, a_{\operatorname{waist}})$ & $(-a_{\operatorname{right}}^{\operatorname{arm}}, -a_{\operatorname{left}}^{\operatorname{arm}}, -a_{\operatorname{right}}^{\operatorname{leg}}, -a_{\operatorname{left}}^{\operatorname{leg}}, -a_{\operatorname{waist}})$\\
        \bottomrule
    \end{tabular}
    \caption{The description and dimensions for each observation and action component.
    The reflection transformation represents the formulation obtained after applying reflection symmetry.
    The height map $H$ is utilized as critic input. 
    }
    \label{tab:obs_and_action_transformation}
\end{table*}

\subsubsection{Symmetry in MDP}

As shown in Fig.~\ref{fig:whole_arch}, the humanoid robots are designed following humanoid morphology, thus exhibiting reflection symmetry in their structure.
Consequently, the MDP exhibits a similar symmetry property described as follows.

As illustrated in Table~\ref{tab:obs_and_action_transformation}, the  \emph{symmetry transformation} denotes the formulation after applying reflection symmetry.
To facilitate reading, we denote the symmetric transformation function of the state, observation, and action as $\mathcal{F}_s$, $\mathcal{F}_o$ and $\mathcal{F}_a$, respectively.
For instance, $o_t$ and $\mathcal{F}_o(o_t)$ represent the robot's observation at $t$-th step before and after reflection symmetry operations, respectively.

Given the MDP $\mathcal{M}$ described in Sec.~\ref{sec:problem_formulation}, we can find that:
\begin{itemize}
\item The transition probability $P$ remains invariant under $\mathcal{F}_o$ and $\mathcal{F}_a$ transformations, i.e. 
\begin{equation}
\label{eq:P_invariance}
    P\left(\mathcal{F}_s\left(s'\right) | \mathcal{F}_s\left(s\right), \mathcal{F}_a\left(a\right)\right) = P\left(s' | s, a\right).
\end{equation}

\item The reward function $R$ is also invariant:
\begin{equation}
\label{eq:R_invariance}
    R\left( \mathcal{F}_s\left(s\right), \mathcal{F}_a\left(a\right)\right) = R\left(s, a\right).
\end{equation}
\end{itemize}
The above invariance properties are derived from the morphological symmetry of the humanoid robot structure.
In other words, the robot takes symmetric actions under symmetric states correspondingly, it will obtain same rewards and reach symmetric states.

Based on Eq.~\eqref{eq:Q_func} and the above formulations, we can obtain the invariance property of critic functions $Q^*$ and $V^*$ of the optimal policy $\pi^*$: 
\begin{equation}
\label{eq:Q_V_invariance}
    Q\left( \mathcal{F}_s\left(s\right), \mathcal{F}_a\left(a\right)\right) = Q\left(s, a\right), \;
    V\left( \mathcal{F}_s\left(s\right)\right) = V\left(s\right).
\end{equation}
This means symmetric states correspond to same values, i.e. same expected cumulative reward in future.
Based on the optimization objective shown in Eq.~\eqref{eq:optimize_pi_objective}, we can derive the symmetry equivariance of the optimal policy $\pi^*$:
\begin{equation}
\label{eq:pi_equivariance}
    \pi^*\left( \mathcal{F}_s\left(s\right)\right) = \mathcal{F}_a \left(\pi^*\left(s\right)\right),
\end{equation}
i.e. the symmetric states correspond to symmetric optimal actions.
This property is consistent with application practice empirically.
For example, when the robot is balanced on its left foot with the right foot unsupported, its optimal action is to lower the right foot. 
Correspondingly, when the robot is in a symmetric state with its left foot unsupported, it needs to take symmetric actions, i.e. lower its left foot.

\subsection{Model Architecture}

As the overall architecture illustrated in Fig.~\ref{fig:whole_arch}, the new method is composed of a history encoder $f_{\operatorname{en}}$, an observation decoder $f_{\operatorname{de}}$, a policy network $f_\pi$ as actor, and a value network $V$ as critic.
This architecture is inspired by DreamWaQ~\cite{nahrendra2023dreamwaq}, which is a state-of-the-art robot locomotion method based on Proximal Policy Gradient (PPO)~\cite{schulman2017proximal}.

\subsubsection{Symmetry Equivariant Actor}

In this work, the actor makes decisions based on temporal observation history $o_{[t-h:t]} = [ o_{t-h} \; o_{t-h+1} \; ... \; o_{t} ]^T$, where $h$ denotes the history length.

As shown in Fig.~\ref{fig:whole_arch}, the \emph{history encoder} $f_{\operatorname{en}}$ inputs history $o_{[t-h:t]}$ and generates latent feature $z$.
In order to train the history encoder $f_{\operatorname{en}}$ to extract appropriate features from the history, an \emph{observation decoder} $f_{\operatorname{de}}$ is utilized to predict next observation $o_{t+1}$, thus is trained though
\begin{equation}
\label{eq:loss_auto_encoder}
    \mathcal{L}_{\operatorname{AE}} =  \operatorname{MSE}(\hat{o}, o_{t+1}), \; \hat{o} = f_{\operatorname{de}}\left(f_{\operatorname{en}}\left(o_{[t-h:t]}\right)\right).
\end{equation}
Afterwards, the \emph{policy network} $f_{\pi}$ makes decisions based on the current observation $o_t$ and the latent feature $z$, i.e. $a_t = f_\pi(o_t, z)$.

\subsubsection{Symmetry Invariant Critic}
As shown in Fig.~\ref{fig:whole_arch}, the critic  evaluates the policy  using the value network $V(H_t, o_t)$, where height map $H$ is privileged information. 
As described in Eq.~\eqref{eq:Q_V_invariance}, the critic is invariant to the symmetric transformation of the input observation.

\subsubsection{Equivariant Neural Network}
In order to incorporate symmetry equivariance described in Eq.~\eqref{eq:pi_equivariance} and Eq.~\eqref{eq:Q_V_invariance}, all networks are constructed utilizing Linear layers and ReLU activations following ESCNN~\cite{cesa2022program}, which implements equivariance based on parameter sharing.
As shown in Fig.~\ref{fig:whole_arch} (bottom right), all equivariant MLPs are designed with symmetry equivariance, given symmetry transformations for inputs and outputs denoted as $\mathcal{F}_{\operatorname{input}}$ and $\mathcal{F}_{\operatorname{output}}$.
Besides, we define the symmetry transformation of latent feature as $\mathcal{F}_z$ shown as follows.
Given latent feature $z$ with even size, i.e. $|z| \operatorname{mod} 2 = 0$, we define $\mathcal{F}_z(z)$:
\begin{equation}
\label{eq:latent_feature_transformation}
    [\mathcal{F}_z(z)]_i =
    \begin{cases}
    z_{i+1} & \text{if } i \text{ is odd}, \\
    z_{i-1} & \text{if } i \text{ is even}.
    \end{cases}
\end{equation}
Based on the symmetry transformation described in Eq.~\eqref{eq:latent_feature_transformation} and Table~\ref{tab:obs_and_action_transformation}, we construct equivariant actor and critic shown in Fig.~\ref{fig:whole_arch}.

\subsection{Training Framework}

The training process is mainly consistent with the standard PPO algorithm.
The critic is trained utilizing MSE loss with ground truth obtained from the trajectory buffer $\mathcal{D}$, i.e. 
$\mathcal{L}_{V} = \operatorname{MSE}(V(H_t, o_t), y)$,
where $y$ is the reward-to-go and is utilized as training labels.

The actor is trained utilizing $\mathcal{L}_{\operatorname{AE}}$ shown in Eq.~\eqref{eq:loss_auto_encoder} and $\mathcal{L}_{\operatorname{PPO}}$ shown as follows:
\begin{equation}
\label{eq:loss_ppo}
\mathcal{L}_{\operatorname{PPO}} = 
\mathbb{E}_{\mathcal{D}}\left[ 
\min\left( 
\rho_{\pi} {A}(s,a), \,
g(\rho_{\pi}) {A}(s,a) 
\right)
\right],
\end{equation}
$\rho_{\pi}=\frac{\pi(a|o, H)}{\pi_{\operatorname{old}}(a|o, H)}$, 
and $g(\rho_{\pi}) = \operatorname{clip}(\rho_{\pi}, 1-\xi, 1+\xi)$. 
$\xi$ is a small hyper-parameter to limit the magnitude of the update, promoting stable and controlled updates.
$A(s,a)$ denotes the advantage of taking $a$ at state $s$, which is obtained utilizing Generalized Advantage Estimation~\cite{schulman2015high}.

The reward functions employed during training are illustrated in Appendix 1.1, which consist of three key components: 
(1) tracking rewards for linear and angular velocity commands;
(2) balance maintenance, such as penalties on velocity on $z$-axis to ensure stability;
(3) regularization terms, such as penalties on action oscillations and torque  exceedance to encourage smooth and reasonable actions.

In order to improve training efficiency and stability, curriculum learning is employed during the training process~\cite{margolis2024rapid}. 
We set different difficulty levels for terrains, task commands, and sensor measurement noise. 
The terrains consist of flat, rough, discrete,
and slope terrains. 
During training, the task difficulty progressively increases as policy performance improves.
In addition, domain randomization is utilized to facilitate the real-world deployment of the policy.
The detailed settings of domain randomization are described in Appendix 1.2.
As shown in Table 2 in the appendix, we implement randomization encompassing ground friction, mass properties, center-of-mass positions, motor parameters, etc.

\section{Experiment}

In this section,  we conduct experiments on the humanoid robot to investigate the following questions:
\begin{enumerate}[label=(\arabic*)]
    \item Can the proposed method be effectively integrated into current DRL framework for humanoid locomotion tasks?
    \item Can the new method generate equivariant policy and improve task performance effectively?
    \item Can the symmetry equivariance property contribute to more stable and coordinated robot motions?
\end{enumerate}

\subsection{Experimental Setup}

In this work, we conduct a series of experiments on the Unitree G1~\cite{unitreeHumanoidRobot}, which is a 27-DoF versatile humanoid robot commonly utilized in robot research.
In this experiment, all the methods are trained utilizing the NVIDIA Isaac Gym simulator~\cite{makoviychuk2021isaacgymhighperformance}.

\begin{table*}[htbp]
    \centering
    \begin{tabular}{lcccc} 
        \toprule 
    \textbf{Metric} & \textbf{DreamWaQ} & \textbf{DreamWaQ-Regu} & \textbf{SE-Policy} & \textbf{SE-Policy (actor only)} \\
    \midrule
    TE-V (cm/s) & $16.43 \pm 9.54$ & $13.91 \pm 8.53$ & $\mathbf{9.85 \pm 1.54}$ & $11.06 \pm 8.63$ \\ 
    Temp-S ($10^{-2}$rad) & $22.52 \pm 2.70$ & $16.58 \pm 2.88$ & $\mathbf{7.86 \pm 1.44}$ & $9.20 \pm 2.19$ \\
    Spat-S ($10^{-2}$rad) & $30.84 \pm 5.201$ & $8.18 \pm 1.46$ & $\mathbf{0.00 \pm 0.00}$ & $0.00 \pm 0.00$ \\
        \bottomrule
    \end{tabular}
    \caption{The experiment results on Tracking Error-Velocity, Temporal Symmetry, and Spatial Symmetry scores.
    Lower values means higher performance.
    The bold scores in the table indicate the optimal results.
    }
    \label{tab:exp_res_quantitative}
\end{table*}

\subsubsection{Baseline Methods}
In this experiment, the following three methods are used as baselines:

\begin{enumerate}[label=(\arabic*)]
    \item \textbf{DreamWaQ}: 
    A state-of-the-art model-free DRL algorithm for legged robot locomotion~\cite{nahrendra2023dreamwaq}. 
    
    \item \textbf{DreamWaQ-Regu}: 
    Enhance DreamWaQ through introducing soft regularization term $\mathcal{L}_{\operatorname{reg}}$ in policy  updates. 
    \begin{equation}
    \label{eq:loss_regu_term}
        \mathcal{L}_{\operatorname{reg}} = 
    \left\|
        \pi\left(\mathcal{F}_o\left(o_{[t-h:t]}\right) \right) - 
        \mathcal{F}_a \left(\pi\left(o_{[t-h:t]}\right)\right)
    \right\|^2.
    \end{equation}
    This auxiliary term penalizes asymmetrical actuation patterns between bilateral joints, which has been widely used in existing works~\cite{abreu2025addressing,ben2025homie,long2024learning,xue2025unified}.
    
    \item \textbf{SE-Policy (actor only)}: 
    Replace critic network of SE-Policy with a vanilla critic based on MLP.
    This ablation study investigates the influence of the critic function's invariance property on humanoid robot performance.
\end{enumerate}

For a fair comparison, all hyper-parameters and environments are kept consistent across various methods.
All methods are trained for over 5K iterations in approximately 4 hours utilizing NVIDIA RTX 4090 GPU devices.
More details of the implementation are given in Appendix 2.1.

\subsubsection{Evaluation Metrics}
In this experiment, the following five quantitative metrics are utilized to evaluate the performance of each method:

\begin{enumerate}[label=(\arabic*)]
\item \textbf{Tracking Error (TE)} evaluates the velocity tracking performance of the policy.
In detail, we analyze the tracking performance from the following three perspective:
\begin{itemize}
    \item \textbf{TE-V}elocity measures the average velocity difference between command and the real velocity vector:
    \begin{equation*}
        \operatorname{TE-V}(\pi) = \mathbb{E}_{\pi}\left[ \left\| v_t - v_{\operatorname{cmd}}\right\| \right].
    \end{equation*}
    \item \textbf{TE-P}osition measures the positional error between the robot's actual and desired positions.
    \item \textbf{TE-O}rientation measures the angle deviation between the robot's actual and directions. 
\end{itemize}

\item \textbf{Temporal Symmetry (Temp-S)} quantifies the difference between joint actions and their symmetric joints' actions in the subsequent half period.
This metric measures the temporal symmetric periodicity during the humanoid locomotion tasks, where lower values corresponding to higher motion coordination.

\item \textbf{Spatial Symmetry (Spat-S)} computes the difference between the action $\pi\left( o_{[t-h:t]}  \right)$ and the symmetric action of symmetric observation, i.e. $\mathcal{F}_a\left( \pi\left( \mathcal{F}_o\left( o_{[t-h:t]} \right)  \right) \right)$.
This metric measures symmetry equivariance property of the policy network directly, where lower values denote higher performance.
\end{enumerate}
More details of these metrics are described in Appendix 2.2.

\begin{figure}[tbp]
    \centering
    \begin{subfigure}[b]{0.450\textwidth}
        \includegraphics[width=\linewidth]{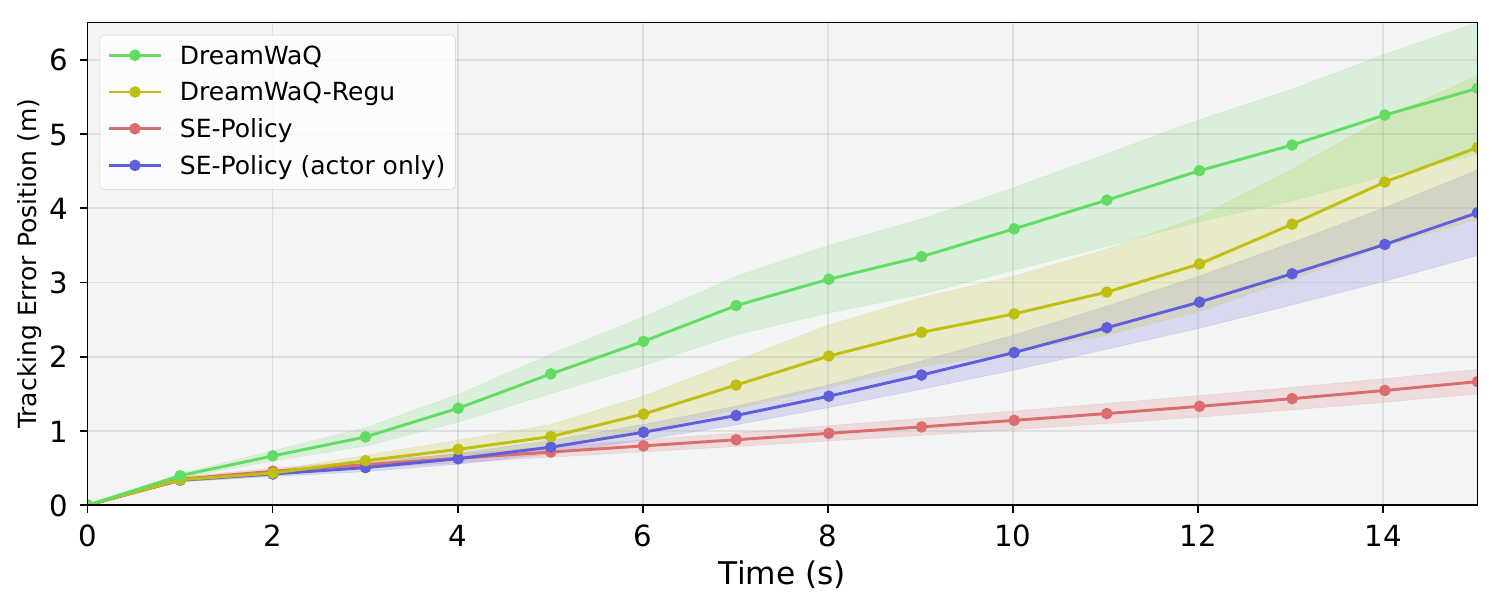} 
        \caption{Tracking Error Position (TE-P)}
        \label{fig:exp_res_TE-P}
    \end{subfigure}
    \begin{subfigure}[b]{0.450\textwidth}
        \includegraphics[width=\linewidth]{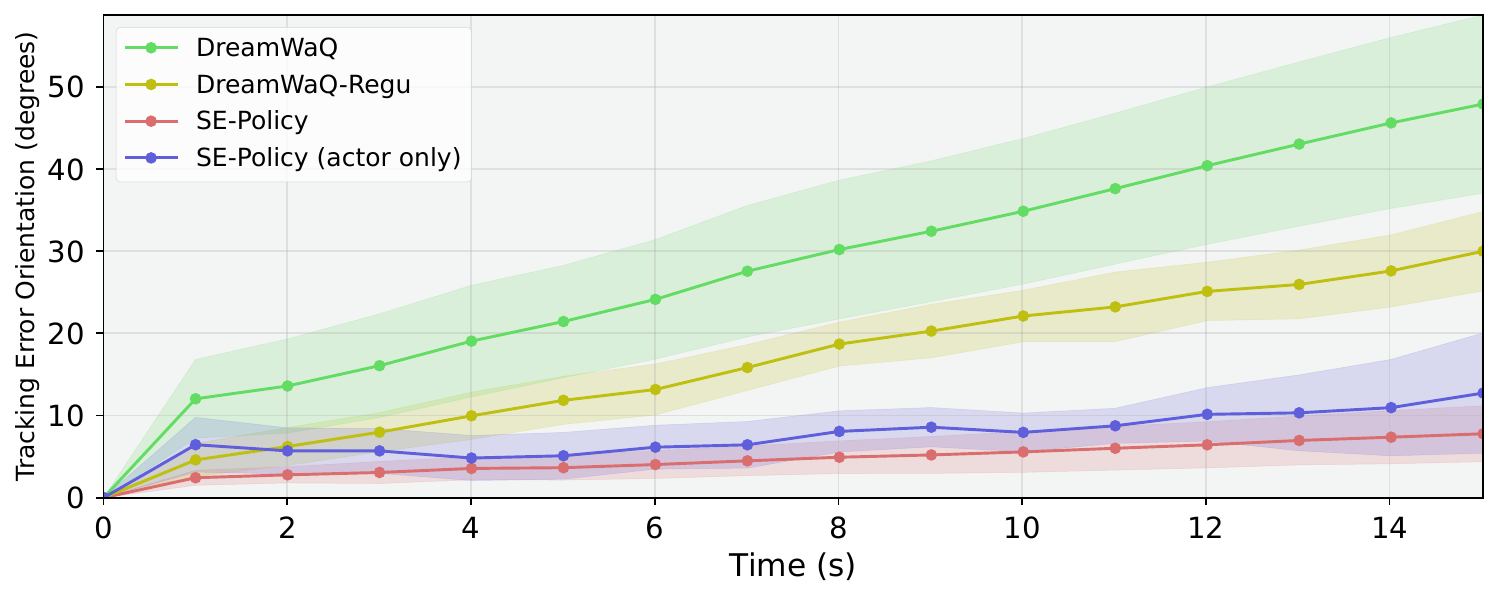}
        \caption{Tracking Error Orientation (TE-O)}
        \label{fig:exp_res_TE-O}
    \end{subfigure}
    \caption{The tracking errors in terms of position (TE-P) and orientation (TE-O) over locomotion time.
    Lines and shadow areas denote mean values and standard errors.
    SE-Policy (red) achieves lower TE-P and TE-O over time than other methods, validating the effectiveness of our method.
    }
    \label{fig:exp_res_TE-P&O}
\end{figure}

\subsection{Experiment Result}

\begin{figure*}[htbp]
    \centering
    \begin{subfigure}[b]{0.233\textwidth}
        \includegraphics[height=\linewidth]{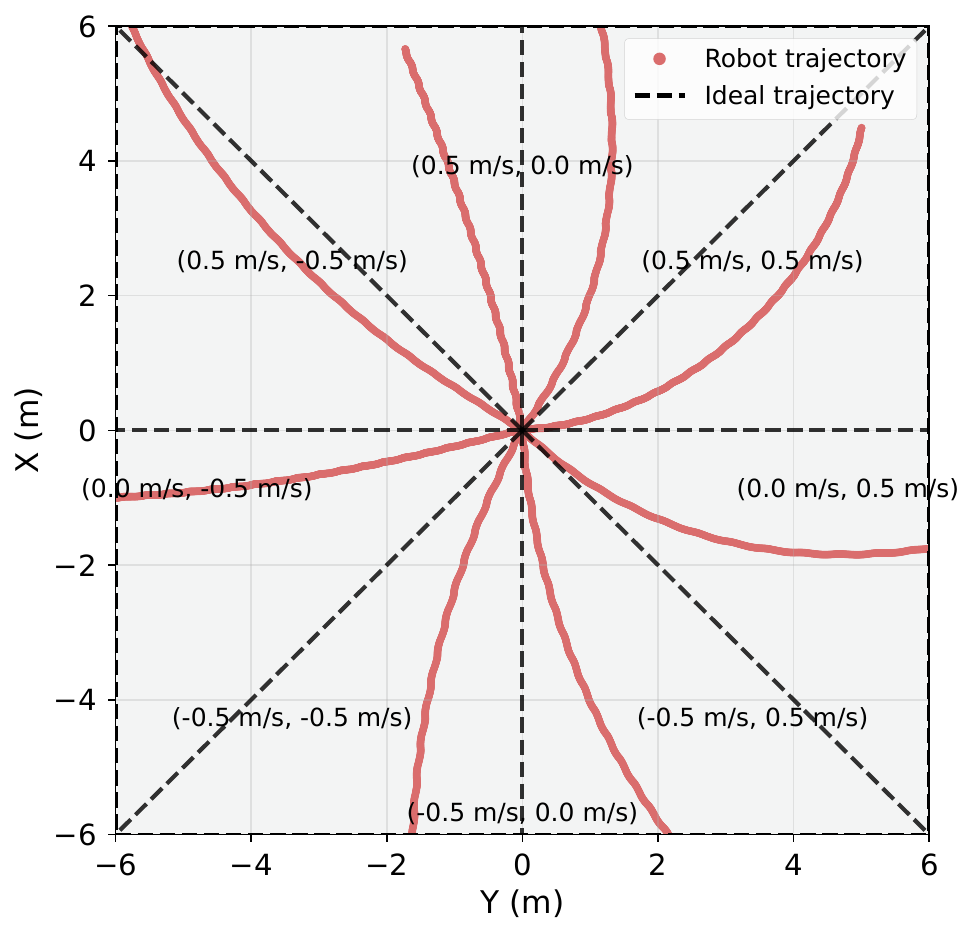}
        \caption{DreamWaQ}
        \label{fig:traj_vis_DreamWaQ}
    \end{subfigure}
    \begin{subfigure}[b]{0.233\textwidth}
        \includegraphics[height=\linewidth]{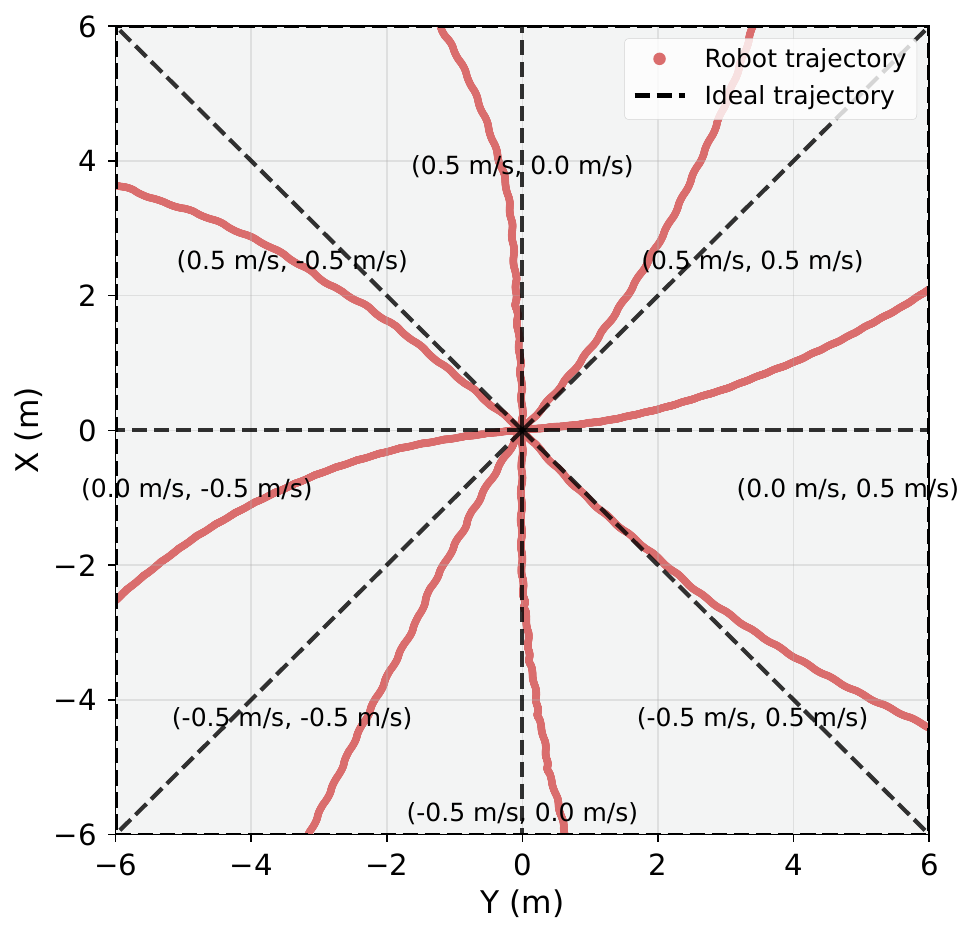}
        \caption{DreamWaQ-Regu}
        \label{fig:traj_vis_DreamWaQ_regu}
    \end{subfigure}
    \begin{subfigure}[b]{0.233\textwidth}
        \includegraphics[height=\linewidth]{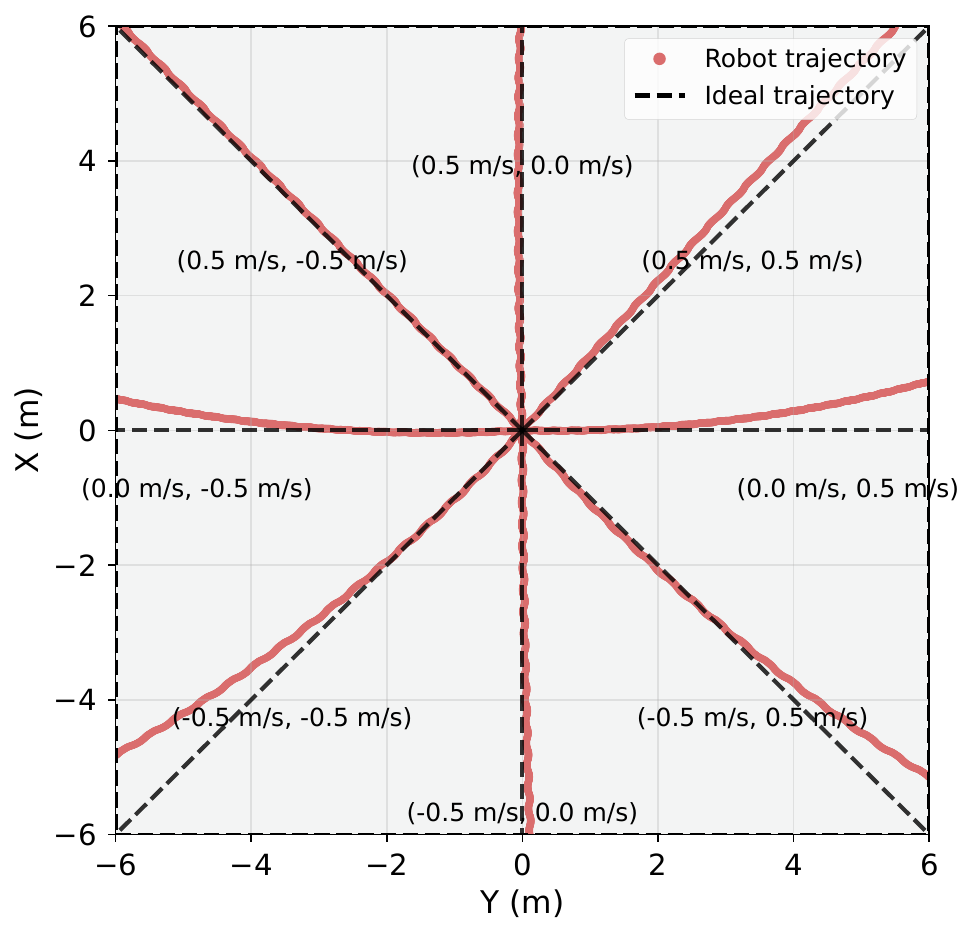} 
        \caption{SE-Policy}
        \label{fig:traj_vis_SE_Policy}
    \end{subfigure}
    \begin{subfigure}[b]{0.233\textwidth}
        \includegraphics[height=\linewidth]{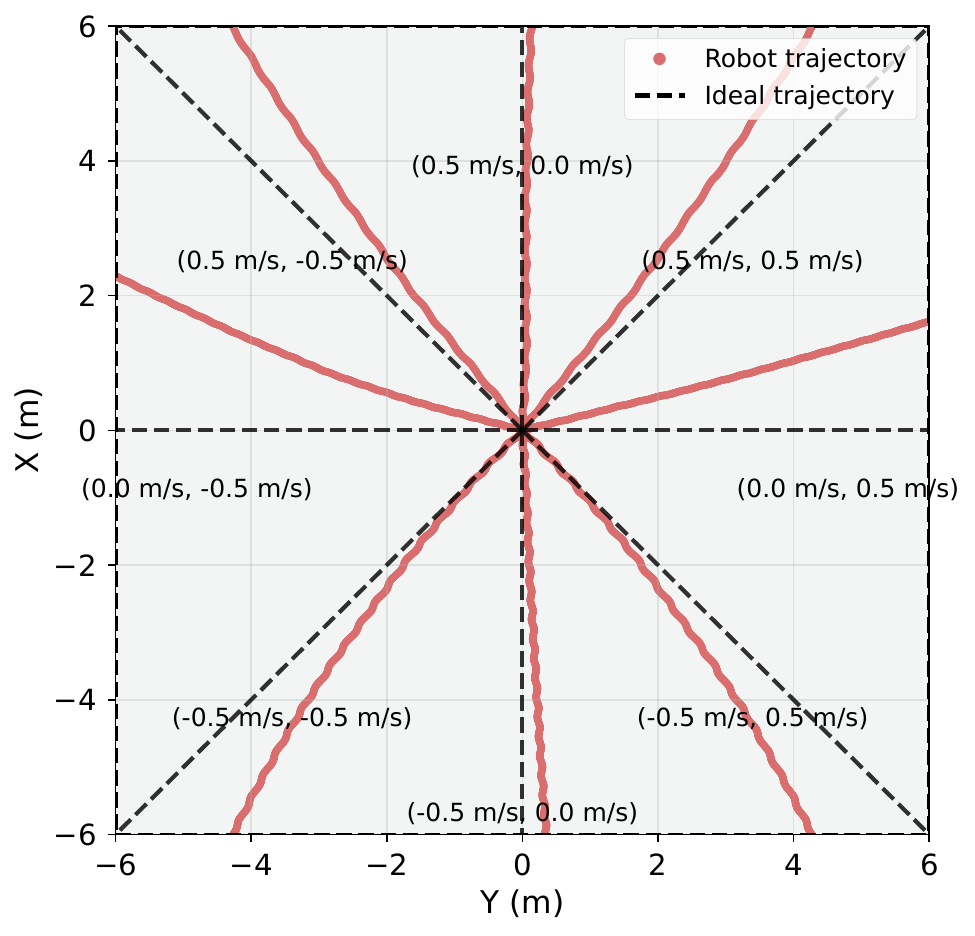}
        \caption{SE-Policy (actor only)}
        \label{fig:traj_vis_SE_Policy_actor}
    \end{subfigure}
    \caption{Visualization of each method's locomotion trajectories.
    The robot is requested to move from the center to eight velocity directions, where dotted lines and red lines are ideal and real trajectories, respectively.
    Our method shown in Fig.~\ref{fig:traj_vis_SE_Policy} outperforms baseline methods on tracking accuracy and trajectory symmetry.
    }
    \label{fig:traj_vis_all}
\end{figure*}

\subsubsection{Task Performance}

In this experiment, the robot is requested to track the random commands in the simulation environment, where velocity commands are randomly sampled with  $|v_x|<0.8 {m/s}$, $|v_y|<0.8 {m/s}$, and $|\omega|<0.5\operatorname{rad/s}$.

As the results shown in Table~\ref{tab:exp_res_quantitative}, our method achieves superior tracking performance.
Specifically, the TE-V of SE-Policy 9.85cm/s is 40.0\% lower than DreamWaQ (16.43 cm/s) and 19.2\% lower than DreamWaQ-Regu (12.19 cm/s), demonstrating the effectiveness of our method.
Besides, the Spat-S of SE-Policy is 0.0, corresponding to the strict symmetry equivariance introduced in the previous section.
The Temp-S of SE-Policy is also lower than other methods, indicating higher symmetric periodicity.
For example, the motion of the left leg in the current phase is more consistent with the right leg in the next phase, which means a more coordinated locomotion style.

The results of Tracking Error-Position (TE-P) and Tracking Error-Orientation (TE-O) are given in Fig.~\ref{fig:exp_res_TE-P&O}.
The x-axis and y-axis denote time and the error value, respectively.
As shown in figures, both TE-P and TE-O increase over time because of the cumulative error.
However, SE-Policy (red line) achieves lower position errors and orientation errors than other methods, because its symmetry equivariance property facilitates more accurate control on velocity and direction.
More details are described in the analysis of the motion visualization.

As illustrated in Table~\ref{tab:exp_res_quantitative} and Fig.~\ref{fig:exp_res_TE-P&O}, DreamWaq-Regu (yellow line) generally achieves higher performance than vanilla DreamWaQ.
For instance, DrwamWaQ-Regu obtains 13.91 TE-V value, 15.3\% lower than vanilla DreamWaQ (16.43), demonstrating the effectiveness of symmetry equivariance achieved by loss regularization shown in Eq.~\eqref{eq:loss_regu_term}.
However, as the Spat-S and Temp-S results shown in Table~\ref{tab:exp_res_quantitative}, the symmetry performance of DrwamWaQ-Regu is weaker than SE-Policy.
This is because the equivariance property of DrwamWaQ-Regu is induced by loss regularization, which is soft and can be violated during inference, corresponding to nonzero Spat-S value in Table~\ref{tab:exp_res_quantitative}.

In addition, despite achieving superior performance over DreamWaQ, SE-Policy (actor only) suffers a non-negligible performance reduction relative to the complete SE-Policy.
For example, SE-Policy (actor only) obtains higher TE-V (12.2\%) and Temp-S (17.0\%) than complete SE-Policy, which means inferior tracking performance and temporal asymmetric motions.
This suggests the necessity of the symmetry invariant critic during the optimization process of equivariant policies in SE-Policy.

\subsubsection{Locomotion Trajectory Visualization}

In this section, in order to give a more intuitive explanation of the difference between SE-Policy and other methods, we visualize the trajectories of each policy in Fig.~\ref{fig:traj_vis_all}.
As shown in the figures, the robot is placed at the center and commanded to move towards eight directions shown as dotted lines, where $v_x, v_y \in \{-0.5m/s, 0m/s, 0.5m/s\}$.
The dotted lines and red lines denote ideal and real trajectories of the robot correspondingly.
The size of the plane is $(12m, 12m)$.

The results illustrated in Fig.~\ref{fig:traj_vis_all} can be analyzed from the following two perspectives:
\begin{enumerate}[label=(\arabic*)]
    \item \textbf{Tracking Accuracy:} As shown in Fig.~\ref{fig:traj_vis_SE_Policy}, robot trajectories of SE-Policy are generally consistent with ideal path (dotted lines), corresponding to the low TE-P and TE-O values described in Fig.~\ref{fig:exp_res_TE-P&O}.
    However, the paths of DreamWaQ in Fig.~\ref{fig:traj_vis_DreamWaQ} deviate from the commanded routes, with cumulative deviation over time, resulting in high tracking errors shown in Table~\ref{tab:exp_res_quantitative}.
    Besides, DreamWaQ-Regu (Fig.~\ref{fig:traj_vis_DreamWaQ_regu}) achieves better performance than DreamWaQ, where the robot stay closer to the planned routes.
    
    \item \textbf{Trajectory Symmetry:} The trajectories in Fig.~\ref{fig:traj_vis_SE_Policy} and Fig.~\ref{fig:traj_vis_SE_Policy_actor} generally exhibit mirror symmetry with respect to the $y=0\operatorname{m}$ line, corresponding to the equivariance property of two policies.
    However, the paths of DreamWaQ-Regu in Fig.~\ref{fig:traj_vis_DreamWaQ_regu} fails to exhibit this result, where paths deviates from commands with a tendency to turn left.
    This is because the equivariance regularization is quite loose, thus the policy cannot be constrained adequately.
\end{enumerate}
Above all, the results in Fig.~\ref{fig:traj_vis_all} demonstrate the effectiveness of strict equivariance induced in SE-Policy, and the necessity of symmetry invariance in critic training.

\subsubsection{Motion Visualization and Analysis}
\label{sec:motion_visulization}

\begin{figure*}[tbp]
    \centering
    \begin{subfigure}[b]{0.47\textwidth}
        \includegraphics[width=\linewidth]{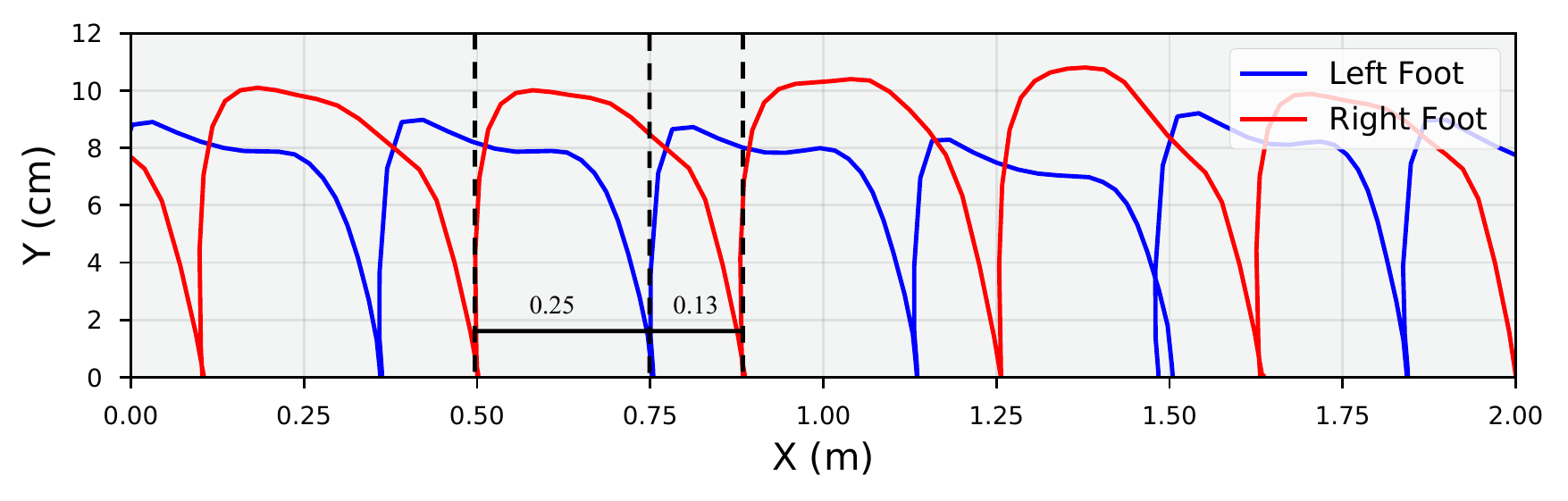}
        \caption{DreamWaQ}
        \label{fig:leg_height_DreamWaQ}
    \end{subfigure}
    \begin{subfigure}[b]{0.47\textwidth}
        \includegraphics[width=\linewidth]{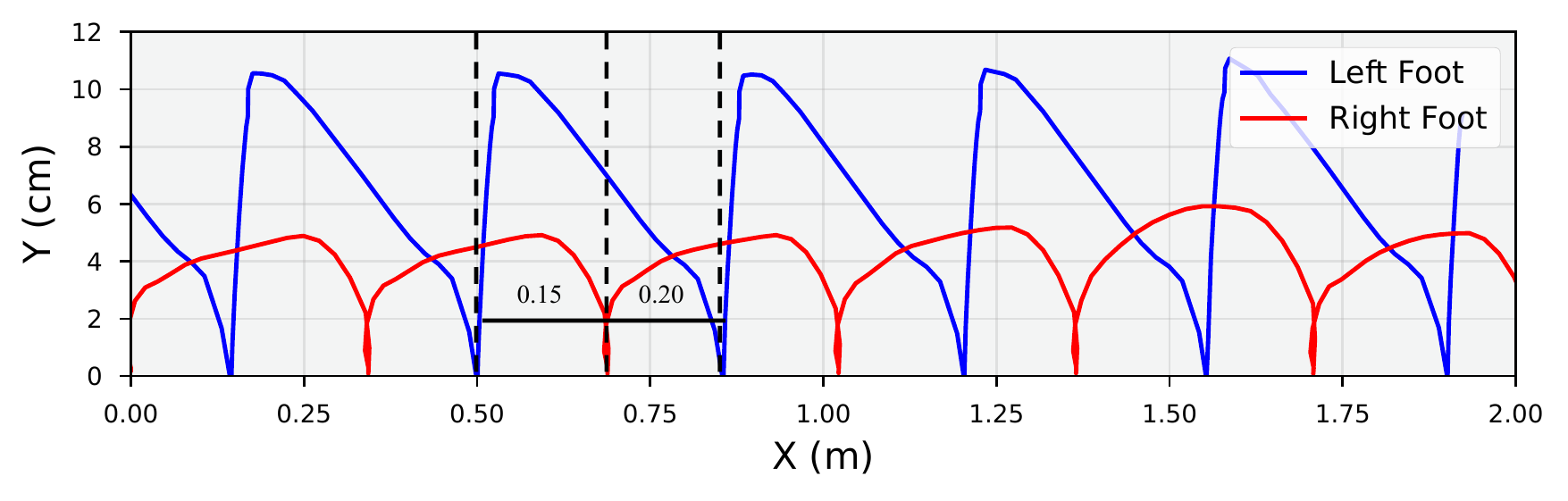}
        \caption{DreamWaQ-Regu}
        \label{fig:leg_height_DreamWaQ_Regu}
    \end{subfigure}

    \begin{subfigure}[b]{0.47\textwidth}
        \includegraphics[width=\linewidth]{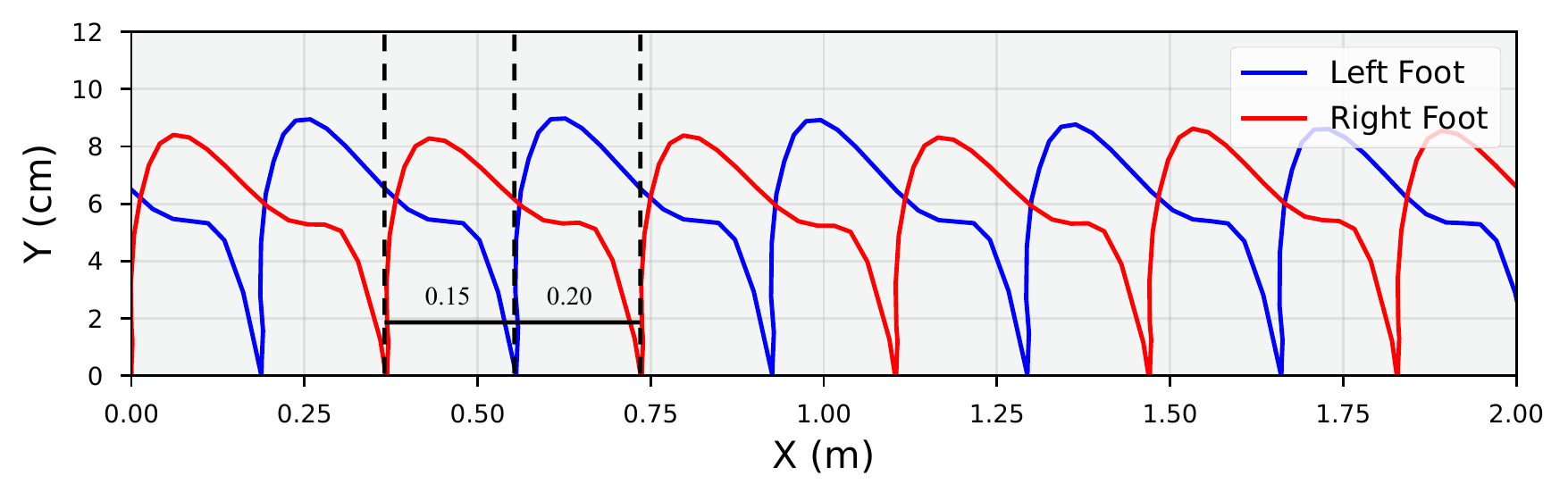} 
        \caption{SE-Policy}
        \label{fig:leg_height_SE-Policy}
    \end{subfigure}
    \begin{subfigure}[b]{0.47\textwidth}
        \includegraphics[width=\linewidth]{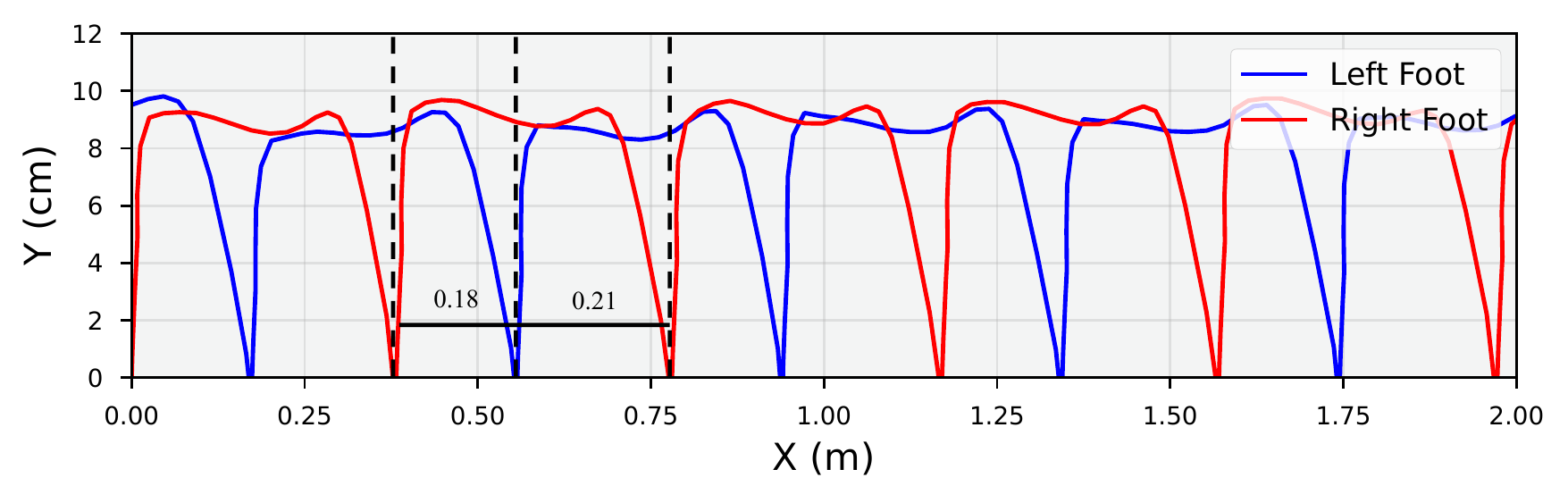}
        \caption{SE-Policy (actor only)}
        \label{fig:leg_height_SE-Policy_actor}
    \end{subfigure}
    \caption{The visualization of foot movement during locomotion on the plane, where $x$-axis denotes moving distance of the robot torso, and and $y$-axis denotes the height of two feet. 
    Our method achieves consistent motions for two feet.
    The motions of two feet generated by SE-Policy are consistent with identical amplitudes, step sizes, and temporal coordination.
    }
    \label{fig:leg_height_all}
\end{figure*}

In this section, we analyze the policy performance through visualizing robot motions.
As described in Fig.~\ref{fig:leg_height_all}, we collect the height of robot feet during locomotion on the plane given constant velocity commands, where two feet are distinguished by two colors.
As shown in Fig.~\ref{fig:leg_height_SE-Policy}, SE-Policy enables two feet to maintain highly periodic motion with nearly constant phase differences, corresponding to the best Temp-S score given in Table~\ref{tab:exp_res_quantitative}.
Besides, two feet are controlled with identical amplitudes and step sizes, resulting in improved coordination stability during robotic movement.

\begin{figure}[ht]
\centering
\begin{subfigure}[b]{0.200\textwidth}
    \includegraphics[width=\linewidth]{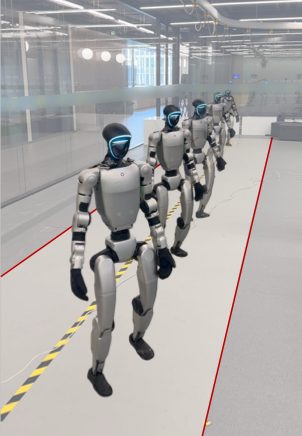}
    \caption{SE-Policy (success)}
    \label{fig:real_robot_show1}
\end{subfigure}
\begin{subfigure}[b]{0.200\textwidth}
    \includegraphics[width=\linewidth]{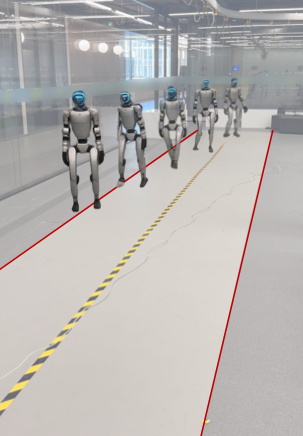}
    \caption{DreamWaQ (fail)}
    \label{fig:real_robot_show2}
\end{subfigure}
\caption{Real world experiments to validate the effectiveness of SE-Policy.
The robot tracks given velocity (warning line) without stepping out of the boundary (red line).
Please refer to the attached video for more details.
}
\label{fig:real_robot_line}
\end{figure}

Additionally, DreamWaQ's movements are quite uncoordinated with inferior symmetric consistency.
As shown in Fig.~\ref{fig:leg_height_DreamWaQ}, there exists non-negligible difference between the motion style of two feet, corresponding to unsatisfying temporal symmetry performance in Table~\ref{tab:exp_res_quantitative}.
The mean step sizes of two feet are $0.13m$ and $0.25m$, leading to uncoordinated motions. 
As shown in Fig.~\ref{fig:leg_height_DreamWaQ_Regu}, DreamWaQ-Regu achieves more consistent step sizes through loss regularization, following the phase signal shown in Table~\ref{tab:obs_and_action_transformation}.
However, there exists significant difference between feet motions on styles and amplitudes, corresponding to inferior Temp-S performance compared to SE-Policy.
This phenomenon suggests that loss regularization described in Eq.~\eqref{eq:loss_regu_term} is effective to improve spatial symmetry, but may cause inconsistent motion styles of symmetric joints, leading to unsatisfying Temp-S performance shown in Table~\ref{tab:exp_res_quantitative}.

\subsection{Real World Experiment}

In this work, we also conduct experiments on the real humanoid robot through Sim-to-real.
We conduct experiment shown in Fig.~\ref{fig:traj_vis_all} on the real robot, where the robot is commanded to track multiple velocity for 10m without stepping out of the boundary (red line).
Part of results are shown in Fig.~\ref{fig:real_robot_line}.
The result is generally consistent with the simulation experiment, where SE-Policy achieves higher tracking accuracy on velocity, position, and orientations, compared to baseline methods.
Besides, we deploy policies on the real robot to evaluate its performance on different terrains, including grass, slope, sand, and stones.
SE-Policy achieves traversability on common unstructured terrains, demonstrating its effectiveness  in real applications.
More experimental results can be found in the supplementary video.

\section{Conclusions}

In this work, we propose Symmetry Equivariant Policy (SE-Policy), a novel DRL method for humanoid robot control tasks.
Inspired by the inherent symmetric morphology of humanoid robots, SE-Policy integrates strict symmetry equivariance and invariance into the actor and critic architecture respectively, without requiring additional hyperparameters during training.
SE-Policy achieves more coordinated and natural robot motions with temporal and spatial symmetry properties.
Multiple experiments are conducted in both simulation and on a real humanoid robot.
The results reveal that SE-Policy consistently outperforms existing methods in terms of tracking accuracy on velocity, position, and orientation, demonstrating the effectiveness of our method.

\section{Acknowledgments}
This work was supported by the National Natural Science Foundation of China (Grant No. 92248303 and No. 62373242), the Shanghai Municipal Science and Technology Major Project (Grant No. 2021SHZDZX0102), and the Fundamental Research Funds for the Central Universities.

\bibliography{aaai2026}

\clearpage
\appendix

\input{Supplementary_Material}

\end{document}

%% file: Supplementary_Material.tex
\section{Implementation Details}


\subsection{Reward Function}

The reward functions utilized in this work are listed in Table~\ref{tab:rewards}, which are mainly composed of the following parts:
\begin{enumerate}[label=(\arabic*)]
    \item The tracking xy and angular velocities reward functions encourage the robot to track velocity commands accurately, including planar (xy) velocities and yaw angular velocity.
    \item The alive reward ensures stability by providing continuous rewards as long as the episode continues without termination.
    \item Penalties are applied to linear z-axis velocity and pitch/roll angular velocities to prevent abnormal oscillations during planar motion.
    \item The orientation and base height rewards promote an upright torso posture at a desired height.
    \item Regularization terms such as action rate and action smoothness constrain abrupt policy output changes, while hip, waist, arm, and torque penalties restrict excessive limb movements for safety.
    \item contact and feet swing height rewards guide the robot’s gait to maintain rhythmic walking patterns.
\end{enumerate}

\begin{table*}[htbp]
    \centering
    \begin{tabular}{llcc} 
        \toprule
        \textbf{Reward terms} & \textbf{Expression} & \textbf{Weight} \\
        
        \midrule
    
       tracking xy velocities & $\exp\left(-\frac{\sum(\text{V}_{lin} - (\text{c}_{x}, \text{c}_{y}))^2}{\sigma}\right)$ & 2.0 \\
    tracking angular velocity & $\exp\left(-\frac{(\omega_{z} - \text{c}_{\omega})^2}{\sigma}\right)$ & 2.0 \\
    alive & $1.0$ & 2.0 \\
    penalize linear velocity on z axis& $\text{v}_z^2$ & -1.0 \\
    penalize  angular velocity on xy axis & $\sum(\omega_{x}^2 + \omega_{y}^2)$ & -0.1 \\
    
    orientation & $\sum\text{g}^2$ & -1.0 \\
    base\_height & $(\text{base\_height} - \text{target\_height})^2$ & -1.0 \\
    
    action\_rate & $\sum(\text{a}_{t} - \text{a}_{t-1})^2$ & -0.005 \\
    action\_smoothness & $\sum(\text{a}_{t} - 2\text{a}_{t-1} + \text{a}_{t-2})^2$ & -0.01 \\

    torques & $\sum\text{torque}^2$ & -1e-5 \\
    hip\_pos & $\sum\theta_{[1,2,7,8]}^2$ & -1.0 \\
    waist\_pos & $\sum\theta_{waist}^2$ & -1.0 \\
    arm\_pos & $\sum(\theta^{arm} - \theta^{arm}_{default})^2$ & -0.1 \\
    
    feet\_swing\_height & $\sum(\text{feet\_pos}_z - 0.03)^2 \cdot \neg\text{contact flag}$ & -20.0 \\
    contact & $\sum\neg(\text{contact} \oplus \text{stance})$ & 1.0 \\
        \bottomrule
    \end{tabular}
    \caption{The description of the reward function utilized in this work.}
    \label{tab:rewards}
\end{table*}

\subsection{Domain Randomization}

The domain randomization settings used in this work are shown in Table~\ref{tab:domain_randomizations}.
\begin{enumerate}[label=(\arabic*)]
    \item The friction term randomizes the friction coefficient of the ground during training, enabling the robot to adapt to terrains of varying materials.
    \item The restitution term randomizes the rebound coefficients during collisions between robot components, simulating real-world elastic collisions across different material surfaces to enhance the policy's robustness against collision dynamics uncertainty.
    \item Randomized mass, center-of-mass position, and moment of inertia settings are used to overcome the sim-to-real gap by ensuring policies trained in simulation generalize to real robots with different mass distributions.
    \item Randomized motor strength and motor offset simulate phase misalignment and imperfect torque output, while randomized kp and kd factors account for discrepancies between control signals and actual torque response. 
    Motor delay mimics signal transmission latency
    These settings are used to match real-world motor behaviors.
\end{enumerate}

Above all, domain randomization techniques are used to align simulated training conditions with real-world dynamics, effectively reducing the sim-to-real gap.

\begin{table}[htbp]
    \centering
    \begin{tabular}{lc} 
        \toprule
        \textbf{Parameters} & \textbf{Range}  \\
        \midrule
        friction & $[0.7, 1.0]$ \\
        restitution & $[0.0, 0.05]$ \\
        base mass & $[-5.0, 5.0]$  \\
        base com & $[-0.015, 0.015]$ \\
        base inertia & $[-0.0005, 0.0005]$ \\
        motor strength & $[0.9, 1.1]$ \\
        motor offset & $[-0.05, 0.05]$ \\
        Kp factor & $[0.9, 1.1]$ \\
        Kd factor & $[0.9, 1.1]$ \\
        motor delay & $[0.02, 0.1]$  \\
        \bottomrule
    \end{tabular}
    \caption{Domain randomization settings in this work.}
    \label{tab:domain_randomizations}
\end{table}

\subsection{Observation and Action}
\begin{figure}[htbp]
\centering
\includegraphics[width=0.99\linewidth]{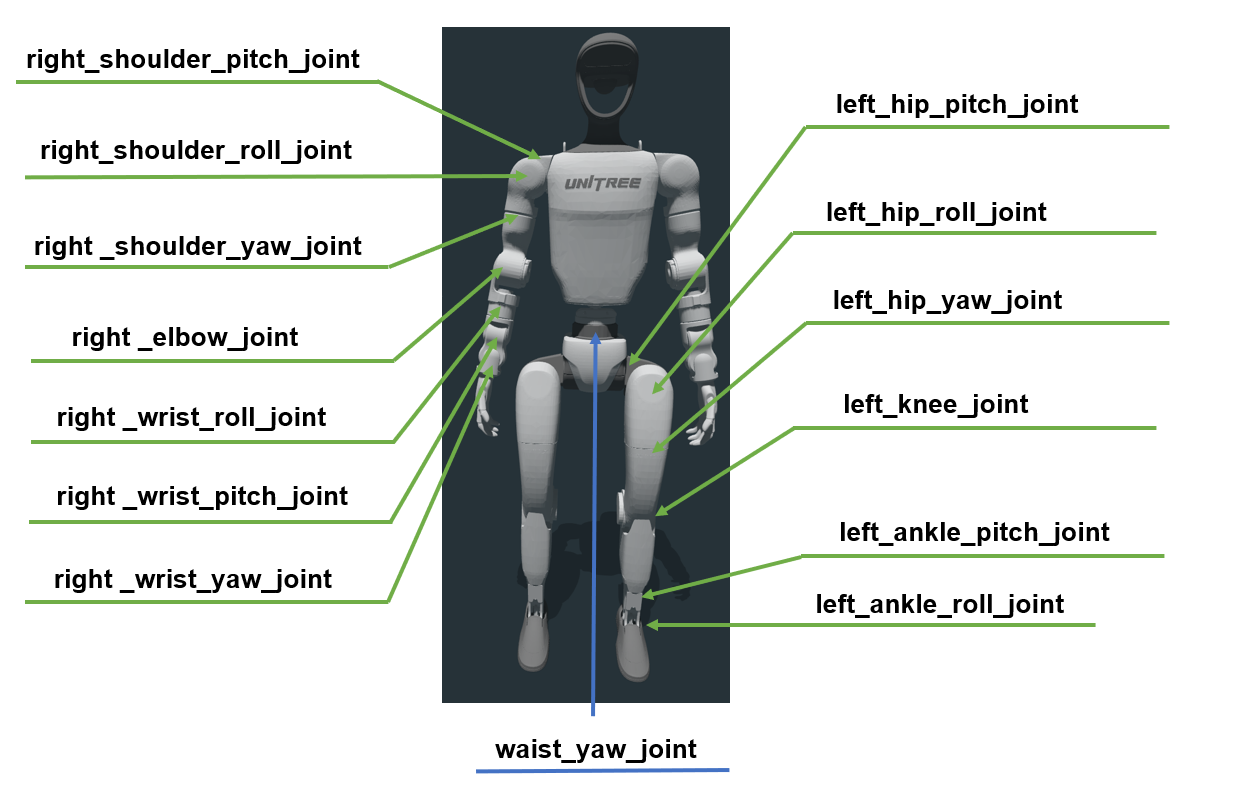} 
\caption{Joint distribution of the G1 robot}
\label{fig:robot_actions}
\end{figure}

In locomotion tasks, the observations obtained from the environment include:
\begin{enumerate}[label=(\arabic*)]
    \item Torso angular velocity of the robot $\omega = (\omega_x, \omega_y, \omega_z)$.
    \item Projection of the gravity vector in the robot's body frame $g=(g_x, g_y, g_z)$.
    \item Velocity commands received by the robot $c=(c_x,c_y,c_{\omega})$ where $c_x,c_y$ represent linear velocity commands on the x and y axes, and $c_{\omega}$ represents angular velocity command on the pitch Euler angle. 
    \item Joint positions of all 27 joints, represented by joint rotation angles. According to the symmetry structure of the robot, the joint position can be sorted into five subsets: $\theta = (\theta_{\operatorname{left}}^{\operatorname{arm}}, \theta_{\operatorname{right}}^{\operatorname{arm}}, \theta_{\operatorname{left}}^{\operatorname{leg}}, \theta_{\operatorname{right}}^{\operatorname{leg}}, \theta_{\operatorname{waist}})$.
    \item Joint velocities of all 27 joints, represented by joint angular velocities $v=(v_{\operatorname{left}}^{\operatorname{arm}}, v_{\operatorname{right}}^{\operatorname{arm}}, v_{\operatorname{left}}^{\operatorname{leg}}, v_{\operatorname{right}}^{\operatorname{leg}}, v_{\operatorname{waist}})$. 
    \item Previous action, represented by the position of the target joint (angles) of the 27 joints $a=(a_{\operatorname{left}}^{\operatorname{arm}}, a_{\operatorname{right}}^{\operatorname{arm}}, a_{\operatorname{left}}^{\operatorname{leg}}, a_{\operatorname{right}}^{\operatorname{leg}}, a_{\operatorname{waist}})$
    \item Phase input, represents the target contact signals for the feet of the robot $\Phi = (\Phi_{sin}, \Phi_{cos})$.
\end{enumerate}
The critic is provided with privileged terrain information Height map $H=(H_{\operatorname{right}}, H_{\operatorname{middle}}, H_{\operatorname{left}})$.
The policy output the target angles of 27-dimensional  joints.
They serve as position control signals for the robot's 27 degrees of freedom (DOF), distributed across its body shown in Fig.~\ref{fig:robot_actions}.
Each upper limb contains three shoulder joints, one elbow joint, and three wrist joints (left and right sides).
Each lower limb consists of three thigh joints, one knee joint, and two ankle joints (left and right sides).
Besides, the robot policy needs to control an additional waist joint.

\section{Experiment Details}

\subsection{Implementation Details}

\begin{table}[htbp]
    \centering
    \begin{tabular}{lc} 
        \toprule
        \textbf{Name} & \textbf{Value}  \\
        \midrule
        Actor network size & $[512, 256, 128]$ \\
        Critic network size & $[512, 256, 128]$ \\
        Encoder network size & $[512, 256, 128, 64]$  \\
        Decoder network size & $[64, 128, 256, 512]$ \\
        Learning rate & $5\times10^{-4}$ \\
        Max grad norm & $1$ \\
        Desired kl & $0.01$ \\
        Activation function & elu \\
        Num mini batches & 4 \\
        Num learning epochs & 5\\
        \bottomrule
    \end{tabular}
    \caption{Network structure and hyperparameter}
    \label{tab:hyper-parameters}
\end{table}

The training pipeline of SE-policy consists of three modules: the actor, the critic, and the auto-encoder (encoder and decoder). 
Their network architectures and detailed training hyperparameter are specified in Table~\ref{tab:hyper-parameters}.
where the max grad norm denotes the maximum gradient value used to normalize the magnitude of network gradients, while the desired kl constrains the degree of change in the policy output distribution after each network update.

\subsection{Evaluation metrics}

The experiment used the following five metrics to evaluate the performance of different methods:

\begin{enumerate}[label=(\arabic*)]
    \item Tracking Error-Velocity (TE-V) : Computes the average difference between the robot's body velocity and the velocity command during movement:
    \begin{equation*}
        \operatorname{TE-V}(\pi) = \mathbb{E}_\pi\left[\left\|(V_x,V_y,\omega_z)-(c_x,c_y,c_{\omega})\right\|\right]
    \end{equation*}

    \item Tracking Error-Orientation (TE-O) : Computes the average difference  between the robot's body orientation $(r,y,p)$ during movement and its ideal orientation $(r_i,y_i,p_i)$ (under perfect velocity command tracking), averaged over N different trajectories.
    \begin{equation*}
        \operatorname{TE-O}(\pi) = \mathbb{E}[\|(r,y,p)-(r_i,y_i,p_i)\|]
    \end{equation*}

    \item Tracking Error-Position (TE-P) : Computes the average difference between the robot's body position $(x,y,z)$ during movement and its ideal position $(x_i,y_i,z_i)$ (under perfect velocity command tracking), averaged over N different trajectories.
    \begin{equation*}
        \operatorname{TE-P}(\pi) = \mathbb{E}_\pi[\|(x,y,z)-(x_i,y_i,z_i)\|]
    \end{equation*}

    \item Temporal Symmetry Score (Temp-S): Compute the average difference between joint actions and their symmetric joints' actions in the subsequent half period, i.e. difference between $a_t$ and $\mathcal{F}_a(a_{t+\frac{1}{2}\delta})$, where $\delta$ is a motion period defined in the reward function.
    \begin{equation*}
        \operatorname{Temp-S}(\pi) = \mathbb{E}_\pi[\|a_t - a_{t+\delta t}\|]
    \end{equation*}

    \item Spatial Symmetry Score (Spat-S) : computes the difference between the action $\pi\left( o_{[t-h:t]}  \right)$ and the action under symmetric observation, i.e. $\mathcal{F}_a\left( \pi\left( \mathcal{F}_o\left( o_{[t-h:t]} \right)  \right) \right)$.
    This metric measures symmetry equivariance property of the policy network directly, where lower values denote higher performance.
    \begin{equation*}
        \operatorname{Spat-S}(\pi) = \mathbb{E}_\pi\left[\| \pi\left( o_{[t-h:t]}  \right) - \mathcal{F}_o\left( o_{[t-h:t]} \right) \|\right]
    \end{equation*}
\end{enumerate}

\subsection{Additional Experiment Results}

\begin{figure}[t]
    \centering
    \begin{subfigure}[b]{0.49\textwidth}
        \includegraphics[width=0.99\linewidth]{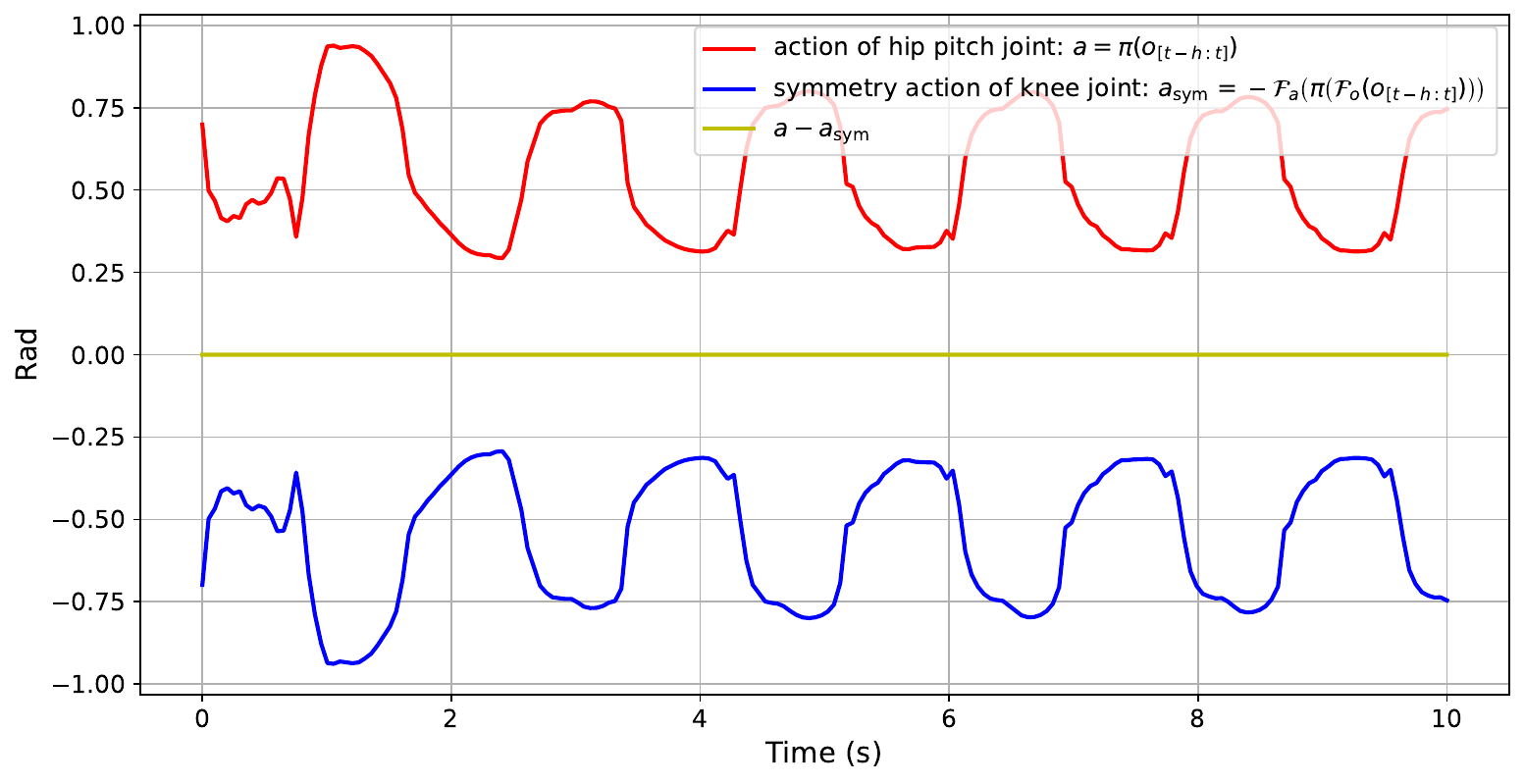} 
\caption{Knee joint}
    \end{subfigure}
    \begin{subfigure}[b]{0.49\textwidth}
        \includegraphics[width=0.99\linewidth]{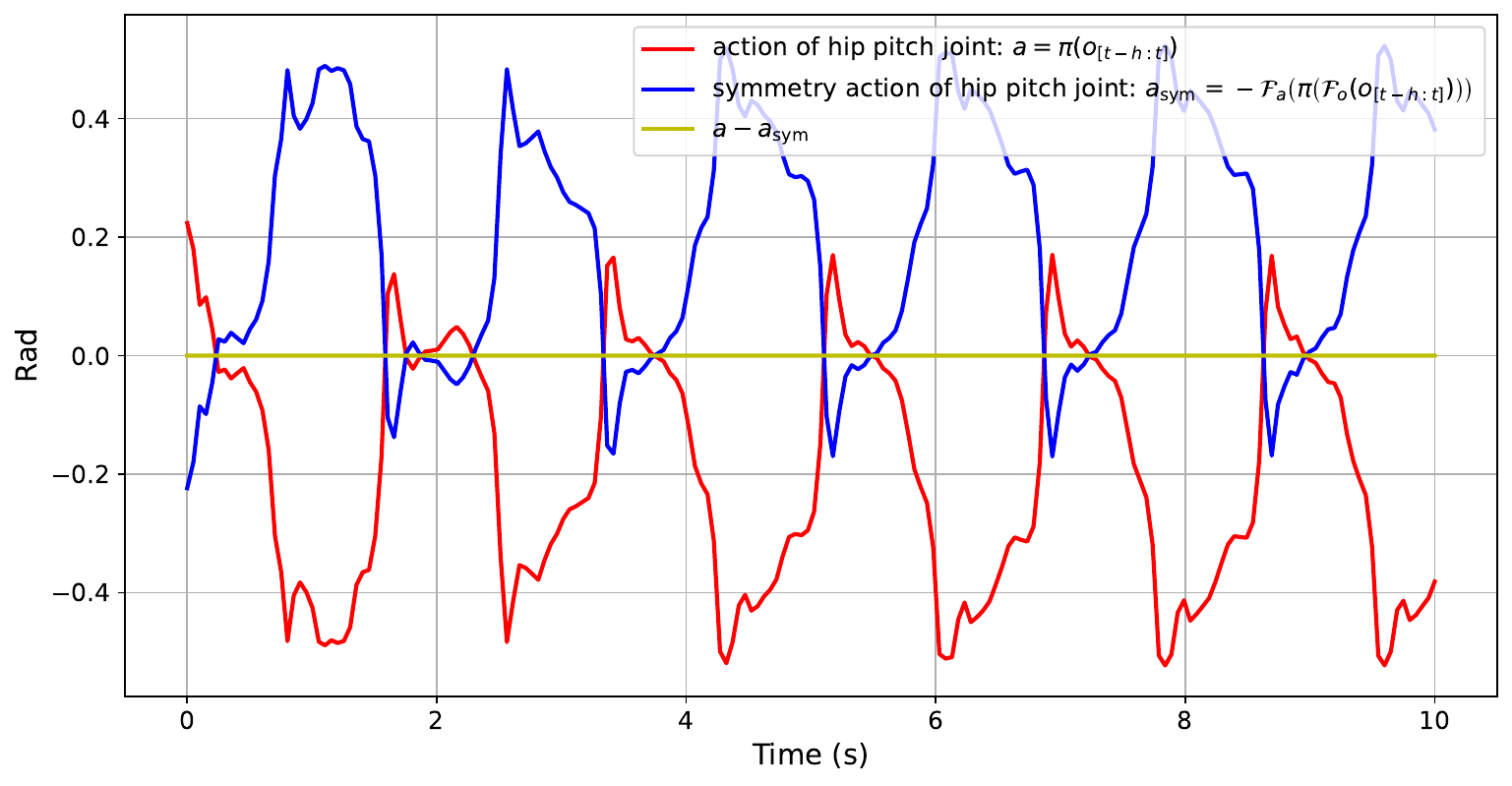} 
\caption{Hip pitch joint}
    \end{subfigure}
    \caption{ 
    The actions $\pi\left( o_{[t-h:t]}\right)$ and ``symmetric actions" $\mathcal{F}_o\left( o_{[t-h:t]} \right)$, i.e. actions under symmetric initial observations of each joint.
    The ``symmetric actions" are negated for clarity.
    These results demonstrate the strict symmetry equivariance of SE-Policy.
    }
    \label{fig:symmetric_actions}
\end{figure}

Besides, we also conduct experiments to verify the strict symmetry equivariance achieved by SE-Policy.
As shown in Fig.~\ref{fig:symmetric_actions}, given observations and symmetric observations, we obtain the corresponding actions $\pi\left( o_{[t-h:t]}\right)$ and ``symmetric actions" $\mathcal{F}_o\left( o_{[t-h:t]} \right)$, shown in red and blue lines correspondingly.
Note that the ``symmetric actions" are negated for clarity.
Both two actions are equal during the locomotion, demonstrating the strict symmetry equivariance obtained in this work.

